%% file: main.tex
\documentclass{article}

\usepackage{spconf,amsmath,graphicx}
\usepackage{amsfonts,amssymb}
\usepackage{algorithm}
\usepackage[noend]{algpseudocode}
\usepackage{caption}
\usepackage{listings}
\usepackage[x11names]{xcolor}
\usepackage{color}
\usepackage{xspace}
\usepackage{dsfont}
\usepackage[percent]{overpic}
\usepackage{xkeyval}
\usepackage{bm}
\usepackage{multirow}
\usepackage[utf8]{inputenc}
\usepackage{mathtools}
\usepackage{hyperref}
\usepackage{lipsum}

\usepackage{array,booktabs,colortbl}
\definecolor{lightgray}{gray}{0.93}
\newcommand{\tablesize}{\scriptsize}
\newcolumntype{C}[1]{>{\centering\arraybackslash}m{#1}}

\makeatletter
\def\BState{\State\hskip-\ALG@thistlm}
\makeatother

\renewcommand{\Re}[1]{\mbox{$\mathds{R}^{#1}$}}
\newcommand{\Ze}[1]{\mbox{$\mathds{Z}^{#1}$}}

\newdimen\origiwspc
\newdimen\origiwstr
\origiwspc=\fontdimen2\font % original inter word space
\origiwstr=\fontdimen3\font % original inter word stretch

\def\eg{\emph{e.g.}}

\def\lone{\mbox{$\ell_1$}}
\def\ltwo{\mbox{$\ell_2$}}

\def\loss{\mbox{$\mathcal{L}$}}

\def\unet{\mbox{U-Net}}

\def\ie{\emph{i.e.\xspace}}

\title{Single Tensor Cell Segmentation using Scalar Field Representations}
\name{{\parbox[c]{0.9\textwidth}{\centering Kevin I. Ruiz Vargas$^{1*}$, Gabriel G. Galdino$^{1*}$, Tsang Ing Ren$^1$, Alexandre L. Cunha$^{2}$\\
  \thanks{\footnotesize\fontdimen2\font=0.35ex
We thank financial support from the Brazilian funding agencies
  FACEPE, CAPES and CNPq (KIRV, GGG, TIR), and from the Beckman Institute at Caltech
    to CIMA -- Center for Image Analysis (ALC).
    $^*$Authors contributed equally. Corresponding author: \texttt{cunha@caltech.edu}.}}}}

\address{
$^1$Centro de Informática, Universidade Federal de Pernambuco, Brazil\\
$^2$Center for Image Analysis, Beckman Institute, California Institute of Technology, USA
}

\begin{document}
\ninept
%\bstctlcite{IEEEexample:BSTcontrol} % This activates the control entry
%\lipsum[4]
%
\maketitle
\input{abstract}

\input{introduction}
%
\input{method}
\input{results}

\input{conclusions}
%
%\newpage
%\balance
%

%\clearpage
\bibliographystyle{IEEEbib}
%\footnotesize
\clearpage
\bibliography{refs}

\end{document}

%% file: abstract.tex
\begin{abstract}
\tolerance=9999
\hyphenpenalty=1000
\exhyphenpenalty=100
We investigate image segmentation of cells under the lens of scalar fields. 
Our goal is to learn a continuous scalar field on image domains such that its segmentation produces robust instances for cells present in images.
This field is a function parameterized by the trained network, and its segmentation is realized by the watershed method.
The fields we experiment with are solutions to the Poisson partial differential equation and a diffusion mimicking the steady-state solution of the heat equation. These solutions are obtained by minimizing just the field residuals, no regularization is needed, providing a robust regression capable of diminishing the  
adverse impacts of outliers in the training data and allowing for sharp cell boundaries.
A single tensor is all that is needed to train a \unet\ thus simplifying implementation, lowering training and inference times, hence reducing energy consumption, and requiring a small memory footprint, all attractive features in edge computing. We present competitive results on public datasets from the literature and show that our novel, simple yet geometrically insightful approach can achieve excellent cell segmentation results.
\end{abstract}

%% file: introduction.tex
\section{Introduction}
\label{sec:intro}
\vspace{-2mm}
%
%\tolerance=9999
%\hyphenpenalty=1000
%\exhyphenpenalty=100
%
Cell instance segmentation is a fundamental step in some areas of quantitative cell biology, \eg\ single cell analysis, yet it remains a challenging task. Difficulties in segmenting include high cell density, complex cell morphology and topology, mixed cell size variability, cell overlaps with indistinct boundaries, limited availability of trustworthy annotated datasets for model training, multimodality of images, and poor image acquisition. All of these contribute to limited results and often require specialized and tuned solutions.

Contemporary deep learning frameworks \cite{ronneberger2015unet,Siddique_2021,prytula2025iaunet,he2017mask, schmidt2018cell, stringer2021cellpose, pachitariu2025cellpose, instanseg, archit2025segment, Cutler2022Omnipose}
% Mask R-CNN \cite{he2017mask}, StarDist \cite{10.1007/978-3-030-00934-2_30},  Cellpose \cite{stringer2021cellpose,pachitariu2025cellpose}, InstaSeg \cite{instanseg} and others 
have significantly advanced segmentation performance by introducing intermediate representation maps that help encode cell shape and boundary information \cite{WEN2022107211,nunes2025survey}. These multi-map approaches have achieved state-of-the-art results while requiring multiple network output heads and elaborate post-processing steps to reconstruct the final instance masks.
%with increasing architectural and computational complexity.

%Our formulation is a two-step process in which a learned model generates a 
%continuous field map for a grayscale image and then segment this field using 
%the classical watershed method to delineate cells. 
%
\begin{figure}[t!]
  \centering
  \includegraphics[width=\columnwidth]{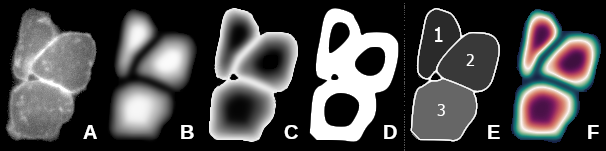}
  \caption{\footnotesize 
    Our proposed segmentation scheme is a two--step process. It initially infers field maps from a single tensor model trained on grayscale images and then segment these maps using the watershed method. In the example above, a cluster of cells (A) admits a Poisson field map (B) generated from our inference model, clearly presenting boundaries separating adjacent cells and background. We invert such map while preserving the background mask to obtain an image (C) suitable for watershed segmentation. Regional {\em h-minima} are then computed (D) and applied as basins by the watershed method to give the instance segmentation in (E) showing boundaries tightly conforming to cell contours. The background mask from (C) eliminates segmented regions lying on it. The field pseudocolormap in (F) helps visualize the Poisson map in (B) and regional maxima extent (dark regions) leading to the compact basins in (D).
  }
    \vspace{-2mm}
    \color{gray}\rule{1.0\columnwidth}{0.2mm}
    \label{fig:pipeline}
    \vspace{-10mm}
\end{figure}
We build upon these works, recent advances in scientific machine learning, and Poisson shape representation developments \cite{gorelick2006shaperef} We propose a straightforward method which casts cell instance segmentation as a single field regression problem combined with the watershed method \cite{beucher1979use}. Instead of predicting multiple intermediate maps, the network learns to estimate a single continuous and smooth scalar field for an entire image domain from which cell instances can be directly extracted. A single field requires just a single tensor thus leading to faster training and inference computing times when compared to multi--map approaches and others. By adopting the watershed method we take advantage of extensive knowledge and successful usage of the method over decades and its geometrical insights.
The crucial step for the watershed to thrive segmenting the cells and background represented by the field is to have the network produce a well behaved, cup--shaped field per cell with deep valleys separating them and a flat background, features we demonstrate with our training (see Fig.\ref{fig:pipeline}).
Upon writing this paper, we discovered that the classical technique in \cite{gardzinski2016human} is similar to ours in concept, but its goal is to try to split existing merged binary masks, a task much simpler from ours and without any learning component.

We explore two realizations of these fields: a Poisson-based representation, in which each cell admits a smooth potential surface obtained by solving the Poisson equation in its domain \cite{gorelick2006shaperef}, and a diffusion-based representation, which mimics the dissipation of continuous heat influx throughout the homogeneous cell until it reaches steady state \cite{hirling2023cell}. Both formulations integrate geometric and boundary cues within a single tensor, eliminating the need for multiple prediction heads and auxiliary maps. 
 
We implement these frameworks using a baseline \unet\ and an enhanced attention-based variant, both trained exclusively with a \lone\ regression loss to predict the scalar field directly from microscope images. Despite their simplicity, these models achieve segmentation accuracy comparable or superior to state-of-the-art multi-map methods while reducing computational cost. The resulting single-channel output provides a mathematically consistent, robust, and efficient representation for instance-level cell segmentation.

% Neural fields have gained much attention recently as a research area on itself. It has shown excellent results, mainly in image synthesis and reconstruction,  scale invariant, and work well in many application areas, from generative to reconstruction models for graphics \cite{park2019deepsdf, xie2022neural}, including image processing \cite{xie2022neural}. These usually employ coordinate--based training to promote scale invariance so scaling is not an issue but regular grids, like images
% PINNs are different as they have a loss which includes residuals of the differential equation and boundary conditions so the networks actually learns to solve the PDE and not only match the sampled ground truth data.

Advances in scientific machine learning, more specifically in physics--informed neural networks, PINNs \cite{raissi2019physics, kharazmi2019variational}, have shown how to parameterize networks to obtain solutions for partial differential equations, PDEs, in a scale independent manner, akin of neural fields \cite{xie2022neural}, and much faster but not necessarily as accurate as established numerical methods (\eg\ finite elements). Training is typically done to minimize residuals of observed data, PDE, and boundary conditions such that learning the physical knowledge modeled by the PDE, the primary goal, is attained. We learned from PINNs that neural networks can indeed predict very well continuous smooth solutions for PDEs but only truly feasible when data residuals are considered \cite{quarteroni2025combining}.
We are thus exclusively interested in the data--driven component to solve the field PDE, disregarding for now any causality principle from the PDE itself. And we solve for the dense image grid, not only for a few domain points, so accuracy, an active research area in PINNs \cite{cuomo2022scientific}, is achieved at its best. 

In the dichotomy between generalist and specialist models, evidence from recent research suggests that generalist models need to specialize to achieve optimal results. It is this need of specialization that questions the true value of generalization, especially when expensive and closed models are adopted. In \cite{gil2024exploring}, the authors demonstrated that generalist models for glioblastoma segmentation (Dice score of 88.98\%) trained on multi-center data were outperformed by specialist models trained on far fewer center-specific data. When generalists were fine-tuned per center they achieved the best results. This illustrates that tuning costly and time-consuming generalist models may often be necessary when specialization is deemed necessary and that specialist models like ours, cheaper to build and maintain, might be sufficient in many cases. The results presented in Table \ref{tab:segmentation_models} reinforce these observations as they demonstrate that our specialist models surpass a proven generalist model in many cases.

% In the dichotomy between generalist and specialist models, evidence from recent research and practice suggests that generalist models need to specialize to achieve optimal results. It is this very specialization need that questions the true value of generalization for developers and end users, especially when expensive and closed models are adopted. In \cite{gil2024exploring}, the authors demonstrated that generalist models for glioblastoma segmentation (Dice score of 88.98\%) trained on multi-center data were outperformed by specialist models trained on far fewer center-specific data. When generalists were fine-tuned per center they achieved the best results. This illustrates the difficulty of building universal models, maybe elusive, and that tuning costly and time-consuming generalist models may often be necessary when specialization and higher accuracy are deemed necessary. It clearly shows the value of specialist models which are cheaper to build and sufficient in many cases. The results presented in Table \ref{tab:segmentation_models} reinforce these observations as they demonstrate that our specialist models surpass a proven generalist model in many cases.

%% file: method.tex
\vspace{-2mm}
\section{Method}
\label{sec:method}
\vspace{-2mm}
\begin{figure}[t!]
  \centering
  \includegraphics[width=\columnwidth]{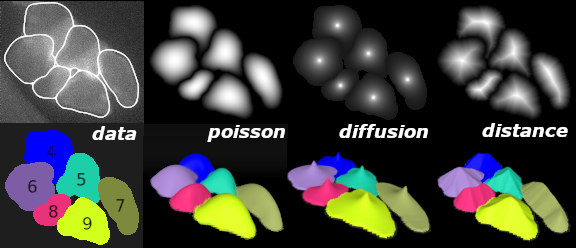}
  \caption{\footnotesize 
    Ground truth field maps $\{U_k\}$ generated for background and each and all cell instances are used in our $\lone$ regression. In this illustration, annotation by a biologist yields labeled cell instances shown in the {\em data} column. For each cell instance their Poisson, diffusion, and Euclidean distance transform field maps, all normalized in $[0,1]$, are shown together with 3D rendering of their respective topographic maps. Background has a field value of zero. The Poisson and diffusion maps are smooth and they have well localized round peaks, offering overall better images and compact seeds for watershed segmentation. 
    %
    %The distance map is the least desired given its non-smoothness and its potential to produce multiple faulty seeds leading to over--segmentation.
    %
    %, specially for non--convex or almost--convex shapes. 
    %The steady--state heat diffusion maps have a rapid decay from their peaks with lower gradients approaching instance boundaries %but are nonetheless sufficient. 
  }
  \vspace{-2mm}
  \color{gray}\rule{1.0\columnwidth}{0.2mm}
    \label{fig:field_maps}
    \vspace{-8mm}
  \end{figure}

{\bf Definitions and Notation}. For a single channel image $I:\Omega\to\Re{}, \Omega\subset\Re{2}$, we assign a ground truth label image $G:\Omega\to\Ze{}$ with $\Omega = \bigcup_{k=0}^{n}\Omega_k$ partitioned into $n$ disjoint subsets $\Omega_k$ with boundaries $\Gamma_k$, each representing a labeled cell mask with centroid pixel $c_k$ and $\Omega_0$ the background region free of cells. We say a pixel at $x\in\Omega$ belongs to $\Omega_k$ if $g(x) = k$, for $g:\Omega\to\Ze{}$ the label function, and $g_k: \Omega\to k$. Similarly, from $G$ we create a ground truth field image $U:\Omega\to\Re{}$ which is built piecewise: $U(x) = u_k(x), x\in\Omega_k$ and $U(x) = 0, x\in\Gamma_k$ where $u_k:\Omega_k\to\Re{}$ is the field function per cell domain $\forall k$, with $u_0(\cdot) = 0$, see Fig.\ref{fig:field_maps}. Thus, the zero level set of $U(\Omega\backslash\Omega_0)$ gives the true segmentation. 
A parameterized network $\zeta_{\theta}(x)$, for $\theta$ network parameters, represents a scalar field $\zeta:\Omega\to\Re{}$. Training using $\{(I_j,U_j)\}_j$ pairs leads to optimized parameters $\theta^*$ such that (abusing the notation) $\zeta_{\theta^*}(I_j)\approx U_j$.
Given a general scalar, continuous field image $U$ the parameterized watershed procedure $w_h(U)\rightarrow\Psi$ segments it into disjoint partitions $\Psi_k$, each a segmentation for a cell, with $\Psi = \bigcup_k\Psi_k$ and for $h\in\Re{}$ the $h-$minima value, see Fig.\ref{fig:pipeline}. Background region $\Psi_0$ is constructed by thresholding $U$ close to zero, $\Psi_0 = \{ x | U(x) < \epsilon, 0 < \epsilon \ll 1 \}$.
Our aim after training is to predict for a general grayscale image $I$ a segmentation $\Psi$ that is close to the true segmentation: $w_h(\zeta_{\theta^*}(I))\rightarrow \Psi\approx\Psi^*$.
We call $\Delta$ the Laplacian operator, $t$ is the time variable, $|r|_p$ the $\ell_p$ norm for vector $r$, $|r|_p = (\sum_i |r_i|^p)^{1/p}$, and $|\Omega|$ the size of set $\Omega$.
\vspace{-2mm}
\subsection{Field Representations for Cells}
\vspace{-1mm}
The heat equation describes how heat $u(x,t)$ propagates over time throughout a continuous medium under an external heat source $f(x,t)$ which for a medium with thermal diffusivity equal to one is
\vspace{-2mm}
\begin{equation}
    \frac{\partial u}{\partial t} = \Delta u + f
    \label{eq:heat}
\end{equation}
Its behavior is governed by $\Delta$ which propagates heat from hot to cold areas of the medium over time. We are particularly interested in the steady state case $\frac{\partial u}{\partial t} = 0$ which is equivalent to solving the Poisson equation $\Delta u = -f$ with given boundary conditions. A Poisson solution can be viewed as the static, steady-state limit of diffusion, describing the equilibrium distribution shaped by boundary and source conditions. Together with diffusion, these physically grounded formulations are continuous potential field representations for instance-level cell segmentation.

Another physical system described by the Poisson equation is the displacement of a thin membrane clamped at its edges, $u(\Gamma) = 0$, under the influence of a uniformly distributed load $f(x)$. We view every cell $\Omega_k$ in a label image $G$ as a membrane clamped at its border $\Gamma_k$ under a uniform load $f(x) = 1$ subject to a displacement field $u_k$. We build $u_k$ individually per cell, normalizing it to $[0,1]$.

We approximate the steady state of eq.\ref{eq:heat} for a cell in two ways: 1) construct a diffusion field for $\partial u/\partial t = 0$ by iteratively spreading influx heat over the cell until stabilization and 2) numerically solve the Poisson equation $\Delta u_k = -1\;\text{in}\;\Omega_k, u(\Gamma_k) = 0$. We call these solutions, respectively, the diffusion and Poisson fields, which, by construction, are deliberately not the same, see Fig.\ref{fig:field_maps}.
\vspace{-2mm}
\subsubsection{\bf Diffusion Regression Targets}
\label{sec:heat}
\vspace{-1mm}
Our diffusion targets are generated by repeatedly injecting constant heat at a cell centroid $c_k$ and applying masked averaging inside it at each iteration until convergence. This process induces a zero-flux (Neumann) boundary during evolution, while fixed border values are imposed only after normalizing the fields $\{u_k\}_k$ in $[0,1]$. As a result, the fields are discretely harmonic away from the source (i.e., $\Delta u_k = 0$ except at $c_k$), producing sharper, steeper peaks than the Poisson potentials. 
%Both approaches produce smooth, morphology-aware scalar maps but arise from distinct boundary conditions and source terms.
%
We maintain a single $u(x)$ to build the entire field image while updating all instances simultaneously using a low pass filter at each iteration $t$:
\vspace{-2mm}
\begin{equation*}
\begin{aligned}
\text{Add 1 to all sources:}\quad
u^{(t+\frac{1}{2})}(x) &= u^{(t)}(x) + \sum_{k} \mathbf{1}\{x=c_k\} \\[-4pt]
\text{Diffusion :}\quad
u^{(t+1)}(x) &= 
\frac{1}{9} \sum_{y\in \mathcal{N}_9(x)} u^{(t+\tfrac{1}{2})}(y)
\end{aligned}
\vspace{-2mm}
\end{equation*}
where $\mathcal{N}_9(x)$ is the $3{\times}3$ grid neighborhood centered at and including $x$. Care is taken to exclude neighbors outside each $\Omega_k$ from the averaging so there is no false contribution from pixels in neighboring cells and from the background $\Omega_0$. We stop iterating for instances $\Omega_k$ when their fields converge, \ie\ $||u_k^{(t+1)} - u_k^{(t)}||_2 < \varepsilon$ ($\varepsilon = 0.01$ in our experiments). This is a unconditionally stable process that maintains a bounded field and converges to a stable morphology-dependent shape while the low pass smoothing suppresses high frequency oscillations \cite{leveque2007finite}.
The total number of iterations depends on the area of cells, the larger the cell the higher the number of iterations to diffuse the heat until steady state. After converging, field values per cell are linearly normalized in $[0,1]$ always guaranteeing $u(\Gamma_k) = 0$ and setting $u(\Omega_0) = 0$. Our diffusion training image is then formed, $u\rightarrow U$.
\begin{figure}[t!]
  \centering
  \includegraphics[width=\columnwidth]{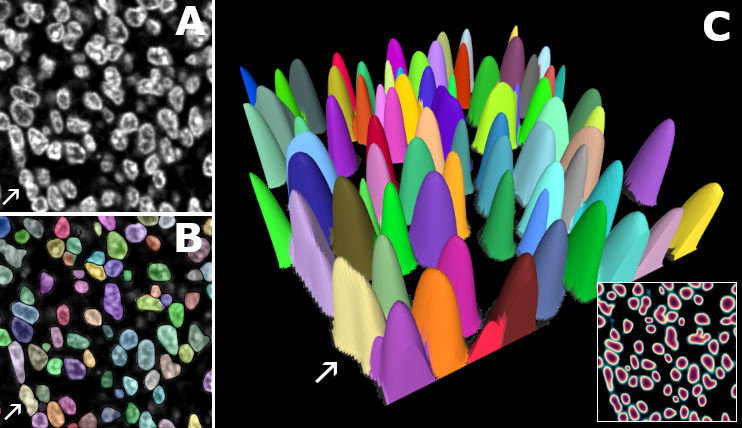}
  \caption{\footnotesize 
    Our Poisson model is capable of segmenting well densely packed and touching nuclei. A small crop of a large tonsil nuclei image segmented using a model trained with T--cells data is shown in (A) with its overlaid segmentation in (B). In panel (C) we show the 3D topographic map of the inferred Poisson field for the crop, colored by instance, with the field depicted in the insert as a 2D pseudocolormap. The field in each nucleus instance is sharply defined, note their peaked shapes in (C), with difficulties only when overlapping is extensive, as in the possible example pointed by arrows.
  }
  \vspace{-2mm}
  \color{gray}\rule{1.0\columnwidth}{0.2mm}
    \label{fig:tonsil}
    \vspace{-8mm}
\end{figure}
\vspace{-2mm}
\subsubsection{\bf Poisson Regression Targets}
\label{sec:poisson}
\vspace{-2mm}

The Poisson formulation defines a unique harmonic field whose values reflect the accumulated influence of all boundary points, yielding a globally coherent field sensitive to cell geometry. One can view $u_k(x)$ as representing the expected time for a random walker starting at $x$ to reach the cell boundary $\Gamma_k$, producing high values near cell centers and lower values towards elongated regions and cell perimeter. Unlike the Euclidean distance transform, which depends only on the nearest boundary point, the Poisson field integrates information over the entire contour, providing a smooth, noise-free representation for complex and branched cell shapes \cite{gorelick2006shaperef} -- see 3D rendering of field in Fig.\ref{fig:tonsil}C.

We numerically solve the Poisson field $u_k(x)$ for a cell $\Omega_k$ using a 
discrete Laplacian over a 5\textendash neighbor stencil in the image grid,
\vspace{-0mm}
\begin{equation}
\Delta u_k(x) = \sum_{y\in\mathcal{N}_4(x)} (u_k(y) - u_k(x)),\quad x,y\in\Omega_k
\end{equation}
\vspace{-0mm}
for $\mathcal{N}_4(x)$ the four immediate neighbors of $x$ in the image grid, always guaranteeing Dirichlet boundary conditions $u(\Gamma_k) = 0$.
This yields a sparse linear system $A\,u_k = -\mathbf{1}$, for $\mathbf{1}$ the unit column matrix, which is efficiently solved using LU factorization. Since $A$ is symmetric positive\textendash definite, the solution is guaranteed to exist and unique for each simply connected region $\Omega_k$. This approach provides a numerically stable and highly accurate solution avoiding the need for iterative refinement or convergence checks. 
Each field $u_k$ is computed independently and linearly normalized to lie in $[0,1]$, being comparable across cells of different sizes and yielding numerically stable regression targets. All $\{u_k\}_k$ are combined in place which together with background $u_0 = u(\Omega_0) = 0$ leads to the field image $U$. This resulting image encapsulates both centrality and shape structure, serving as a single\textendash channel regression target for network training.

We trained our field regression models using an \lone\ loss -- Mean Absolute Error -- which corresponds to the sum of $\ell_{1}$ norm of residuals $r(x) = \hat u(x) - u(x), x\in\Omega$, between the predicted $\hat{u}(x)$ and ground truth $u(x)$ fields
\begin{equation}
\loss = \frac{1}{|\Omega|}\sum_{x\in\Omega} |r(x)|_1 = \frac{1}{|\Omega|} \sum_{x\in\Omega} \big| \hat{u}(x) - u(x) \big|_1
\label{eq:loss}
\end{equation}
Minimizing \loss\ leads to an optimal $u^*$, $u^* = \arg\min_{\hat{u}} \loss$, our field predicton.
We opted for \lone\ norm given its advantages over the commonly used \ltwo\ norm. Of importance, \lone\ gives a sharper field enhancing the separation between adjacent and touching cells when compared to \ltwo, as we have evidenced during development. This is a well known feature established in previous works, \eg\ \cite{chan2005aspects}. Also desirable, using \lone\ makes the optimization less sensitive to outliers potentially present in learning data.  Using \ltwo\ is akin of Gaussian smoothing which smears the signal making difficult to obtain a more precise cell boundary. The \lone\ norm is definitely a better option to achieve sharp boundaries and more reliable instance separation.
\vspace{-2mm}
\subsubsection{\bf Watershed Segmentation}
\vspace{-1mm}
Field predictions are good candidates for segmenting using the watershed method because they are firmly bounded in $[0,1]$ per cell, regardless of cell size, and they have smooth dome--like landscapes with peak values that, when inverted, are excellent compact basins. We have two parameters that guide the watershed segmentation: \mbox{$0<\epsilon\ll1$}: responsible for defining via thresholding the background region in the prediction image (recall $\Psi_0$ in the beginning of this section); and \mbox{$h\in(0,1]$}: it is needed to define the basins via $h-$minima. 
Large $\epsilon$ values underestimate the true size of cells and too tiny values overestimate cell size. In our experiments we have used with success $\epsilon\in[0.05,0.10]$, mostly $\epsilon = 0.05$ showing how good the predictions are when capturing cell boundaries. A value of $h\in[0.05,0.40]$ has worked well. These parameters have been the least of our concerns, no tuning really necessary, attesting again the good quality of our predictions.
\begin{figure}[b!]
  \centering
  \includegraphics[width=\columnwidth]{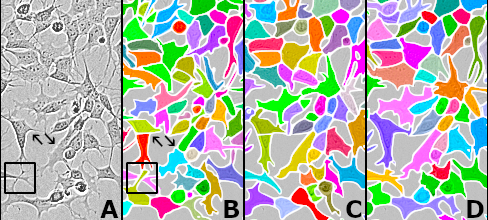}
  \caption{\footnotesize 
    Test segmentation for SH-SHY5Y label-free cells from LIVECell dataset \cite{edlund2021livecell}. We enhanced the original phase-contrast image to help visualization of its cells (A) and their segmentations -- manual ground truth (B), diffusion (C), and Poisson (D). All methods have relatively low score (see table \ref{tab:segmentation_models}) partially due to questionable ground truth (see arrows), overlaps, and crossovers as illustrated by squares in (A) and (B). Nonetheless, we show here that our method is able to segment cells imaged with challenging phase-contrast signals.
  }
  \vspace{-2mm}
  \color{white}\rule{1.0\columnwidth}{0.2mm}
    \label{fig:livecell}
    \vspace{-8mm}
\end{figure}
\vspace{-2mm}
\subsubsection{\bf Training Details}
\vspace{-2mm}
Two neural network modalities were employed to regress the scalar fields for the diffusion and Poisson  representations. For the diffusion formulation, a baseline U-Net was used.
We implemented an enhanced MA-UNet for the Poisson formulation that integrates residual connections, attention modules, and a compact Transformer encoder \cite{cai2022ma}. Each encoder stage uses residual blocks with Group Normalization and SiLU activations, stabilized by a spatial dropout layer ($p = 0.3$). At the bottleneck, a Transformer encoder with eight attention heads and three layers captures long-range dependencies, while dual attention modules in the decoder refine feature fusion and emphasize boundary regions. The network reaches a bottleneck width of 1536 channels, providing high representational capacity and enabling detailed reconstruction of local structures within the Poisson field.
This Poisson formulation was trained with AdamW optimizer ($10^{-4}$ learning rate, $10^{-5}$ weight decay) and a Cosine Annealing with Warmup learning-rate schedule consisting of a 10-epoch warmup followed by cosine decay to 1\% of the initial rate over 200 training epochs for each dataset. 

%% file: results.tex
\vspace{-4mm}
\section{Results}
\vspace{-2mm}
\label{sec:results}
\begin{figure}[t!]
    \centering
        \includegraphics[width=.99\linewidth]{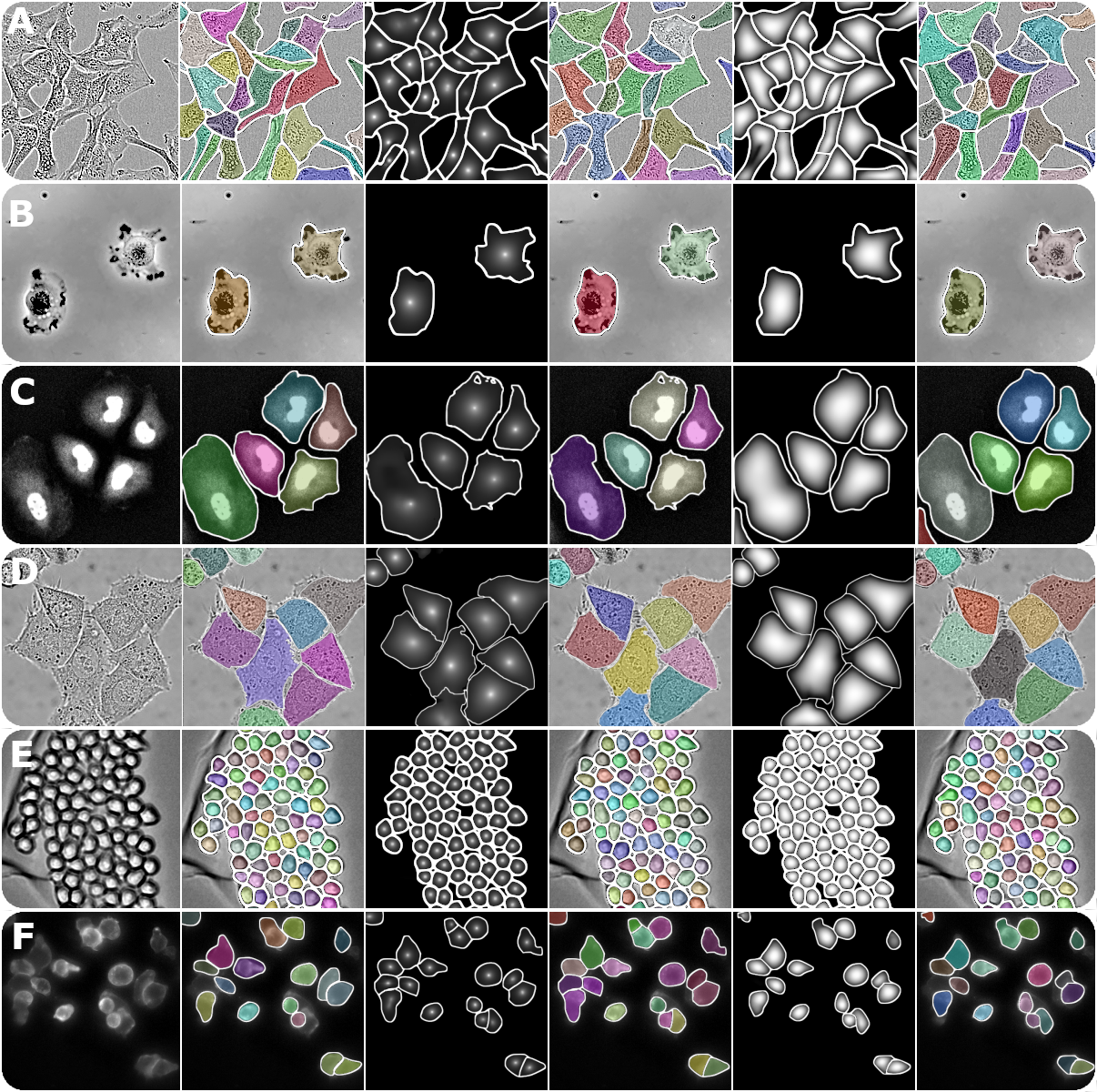}
    \caption{\footnotesize Example segmentations for datasets in Table.\ref{tab:segmentation_models}. Each panel contains, from left to right, small crop of original image, ground truth, Diffusion map and corresponding overlaid colored segmentation, and Poisson map and its segmentation. Datasets are A-172 (A), PhC-C2DH-U373 (B, not on Table.\ref{tab:segmentation_models}), Fluo-C2DL-Huh7 (C), DIC-C2DH-HeLa (D), BF-C2DL-HSC (E), and T-Cells (F). The models segment well packed cells and in diverse modalities.
    }
    \vspace{-2mm}
    {\color{gray}\rule{0.999\linewidth}{0.2mm}}
    \vspace{-9mm}
    \label{fig:results}
\end{figure}

We have trained and tested our networks using publically available datasets \cite{mavska2023cell,edlund2021livecell} and a private T--cells dataset \cite{guerrero2018multiclass} using advocated data partitions, when available, for training, validation, and testing, see Table\ref{tab:segmentation_models}.
%The number of images used for each split (training:validation:testing) is indicated in brackets below each dataset name in Table~\ref{tab:segmentation_models}.
This selection prioritizes diversity in three aspects: (i) imaging modality (phase-contrast, fluorescence, DIC, bright-field), (ii) cell size, morphology and density, including crowded scenes with touching instances, and (iii) variation in dataset scale, ranging from few images to large collections with thousands.
We adopted popular metrics IoU, Dice coefficient, and Panoptic Segmentation Quality to evaluate our models and a well established generalist model \cite{pachitariu2025cellpose}. Results for these testing images show that our method generates segmentations on par with or generally superior to the generalist model. We have not fine--tuned the generalist model to possibly improve it. Figures 3-5 demonstrate small snips of segmentations obtained with our models and ground truth for comparison.

\begin{table}[t!]
\footnotesize
\centering
\tablesize
\setlength{\tabcolsep}{7pt}
\renewcommand{\arraystretch}{1.1}
\newcolumntype{s}{>{\columncolor{lightgray}}c}
\begin{tabular}{|C{2.0cm}|c|c|c|c|}
\hline
\textbf{Dataset} & \textbf{Metric} & \textbf{Diffusion} & \textbf{Poisson} & \textbf{Cellpose-SAM} \\
\hline

\multirow{5}{*}{\shortstack{BF\mbox{-}C2DL\mbox{-}HSC\\{\footnotesize[\,2469:529:530\,]}}}
 & IoU & 0.9357 & \textbf{0.9419} & 0.8223 \\
 & \cellcolor{lightgray}Dice & \cellcolor{lightgray}{0.9664} & \cellcolor{lightgray}\textbf{0.9698} & \cellcolor{lightgray}{0.9014} \\
 & PQ & \textbf{0.9691} & 0.9184 & 0.8341 \\
 & \cellcolor{lightgray}SQ & \cellcolor{lightgray}{0.9699} & \cellcolor{lightgray}\textbf{0.9739} & \cellcolor{lightgray}{0.9014} \\
 & RQ & \textbf{0.9991} & 0.9416 & 0.9003 \\ \hline

\multirow{5}{*}{\shortstack{DIC\mbox{-}C2DH\mbox{-}HeLa\\{\footnotesize[\,117:26:25\,]}}}
 & IoU & 0.8647 & \textbf{0.8945} & 0.7697 \\
 & \cellcolor{lightgray}Dice & \cellcolor{lightgray}{0.9247} & \cellcolor{lightgray}{\textbf{0.9431}} & \cellcolor{lightgray}{0.8649} \\
 & PQ & 0.8533 & \textbf{0.8874} & 0.6706 \\
 & \cellcolor{lightgray}SQ & \cellcolor{lightgray}{0.8717} & \cellcolor{lightgray}\textbf{0.8948} & \cellcolor{lightgray}{0.7077} \\
 & RQ & 0.9783 & \textbf{0.9917} & 0.9137 \\ \hline

\multirow{5}{*}{\shortstack{Fluo\mbox{-}C2DL\mbox{-}Huh7\\{\footnotesize[\,9:2:2\,]}}}
 & IoU & 0.7518 & \textbf{0.7756} & 0.7373 \\
 & \cellcolor{lightgray}Dice & \cellcolor{lightgray}{0.8401} & \cellcolor{lightgray}\textbf{0.8597} & \cellcolor{lightgray}{0.8338} \\
 & PQ & 0.6302 & \textbf{0.8085} & 0.7680 \\
 & \cellcolor{lightgray}SQ & \cellcolor{lightgray}{0.8392} & \cellcolor{lightgray}\textbf{0.8744} & \cellcolor{lightgray}{0.8300} \\
 & RQ & 0.7482 & \textbf{0.9200} & 0.9198 \\ \hline

\multirow{5}{*}{\shortstack{PhC\mbox{-}C2DL\mbox{-}PSC\\{\footnotesize[\,420:90:90\,]}}}
 & IoU & \textbf{0.8931} & 0.8858 & 0.8140 \\
 & \cellcolor{lightgray}Dice & \cellcolor{lightgray}\textbf{0.9411} & \cellcolor{lightgray}{0.9362} & \cellcolor{lightgray}{0.8940} \\
 & PQ & \textbf{0.9342} & 0.9161 & 0.8530 \\
 & \cellcolor{lightgray}SQ & \cellcolor{lightgray}\textbf{0.9482} & \cellcolor{lightgray}{0.9366} & \cellcolor{lightgray}{0.8989} \\
 & RQ & \textbf{0.9845} & 0.9767 & 0.9441 \\ \hline

\multirow{5}{*}{\shortstack{Fluo\mbox{-}N2DH\mbox{-}GOWT1\\{\footnotesize[\,128:28:28\,]}}}
 & IoU & 0.9362 & \textbf{0.9617} & 0.9479 \\
 & \cellcolor{lightgray}Dice & \cellcolor{lightgray}{0.9663} & \cellcolor{lightgray}{\textbf{0.9803}} & \cellcolor{lightgray}{0.9723} \\
 & PQ & \textbf{0.9516} & 0.8682 & 0.9505 \\
 & \cellcolor{lightgray}SQ & \cellcolor{lightgray}{0.9658} & \cellcolor{lightgray}{0.9520} & \cellcolor{lightgray}\textbf{0.9716} \\
 & RQ & \textbf{0.9850} & 0.8936 & 0.9777 \\ \hline

\multirow{5}{*}{\shortstack{PhC\mbox{-}C2DH\mbox{-}U373\\{\footnotesize[\,161:35:34\,]}}}
 & IoU & {\bf 0.9496} & 0.9454 & 0.9046 \\
 & \cellcolor{lightgray}Dice & \cellcolor{lightgray}\textbf{0.9733} & \cellcolor{lightgray}{0.9705} & \cellcolor{lightgray}{0.9428} \\
 & PQ & \textbf{0.9573} & 0.9303 & 0.9046 \\
 & \cellcolor{lightgray}SQ & \cellcolor{lightgray}\textbf{0.9719} & \cellcolor{lightgray}{0.9692} & \cellcolor{lightgray}{0.9568} \\
 & RQ & \textbf{0.9843} & 0.9584 & 0.9431 \\ \hline

\multirow{5}{*}{\shortstack{T\mbox{-}cells\\{\footnotesize[\,77:17:17\,]}}}
 & IoU & 0.7597 & \textbf{0.7964} & 0.7699 \\
 & \cellcolor{lightgray}Dice & \cellcolor{lightgray}{0.8529} & \cellcolor{lightgray}{\textbf{0.8823}} & \cellcolor{lightgray}{0.8632} \\
 & PQ & 0.8543 & \textbf{0.8952} & 0.8013 \\
 & \cellcolor{lightgray}SQ & \cellcolor{lightgray}{0.8877} & \cellcolor{lightgray}\textbf{0.9100} & \cellcolor{lightgray}{0.8446} \\
 & RQ & 0.9678 & \textbf{0.9819} & 0.8858 \\ \hline

\multirow{5}{*}{\shortstack{A172\\{\footnotesize[\,388:68:152\,]}}}
 & IoU & 0.6380 & 0.6796 & \textbf{0.7226} \\
 & \cellcolor{lightgray}Dice & \cellcolor{lightgray}{0.7596} & \cellcolor{lightgray}{0.7936} & \cellcolor{lightgray}{\textbf{0.8267}} \\
 & PQ & 0.6416 & \textbf{0.6794} & 0.6408 \\
 & \cellcolor{lightgray}SQ & \cellcolor{lightgray}{0.7491} & \cellcolor{lightgray}\textbf{0.7652} & \cellcolor{lightgray}{0.7149} \\
 & RQ & 0.7809 & \textbf{0.8202} & 0.7753 \\ \hline

\multirow{5}{*}{\shortstack{SH\mbox{-}SY5Y\\{\footnotesize[\,449:79:176\,]}}}
 & IoU & 0.5271 & 0.5892 & \textbf{0.6484} \\
 & \cellcolor{lightgray}Dice & \cellcolor{lightgray}{0.6693} & \cellcolor{lightgray}{0.7236} & \cellcolor{lightgray}{\textbf{0.7736}} \\
 & PQ & 0.5635 & \textbf{0.6236} & 0.6019 \\
 & \cellcolor{lightgray}SQ & \cellcolor{lightgray}{0.7342} & \cellcolor{lightgray}\textbf{0.7529} & \cellcolor{lightgray}{0.7153} \\
 & RQ & 0.7267 & \textbf{0.7973} & 0.7952 \\ \hline
\end{tabular}
\label{tab:segmentation_models}
\vspace{-2mm}
\caption{\footnotesize 
Quantitative comparison of the Diffusion, Poisson, and Cellpose-SAM models across diverse microscopy datasets. For each dataset, brackets show number of images used for [training:validation:testing]. Mean metric values reported for testing images include IoU, Dice coefficient, Panoptic Quality (PQ), Segmentation Quality (SQ), and Recognition Quality (RQ), \mbox{PQ = SQ x RQ}, computed using {\tt TorchMetrics} package. Top scores for all datasets and metrics amount to 31\% for Diffusion, 58\% for Poisson, and 11\% for Cellpose-SAM, showing that our models, with 89\% top scores, are superior to a superhuman state-of-the-art model.  
}
\label{tab:segmentation_models}
\vspace{-1mm}
\color{gray}\rule{0.99\columnwidth}{0.2mm}
\vspace{-4mm}
\end{table}

%% file: conclusions.tex
\vspace{-4mm}
\section{Conclusions}
\label{sec:conclusions}
\vspace{-2mm}
We investigated steady-state diffusion and static Poisson representations as unified mechanisms for learning continuous descriptions of cellular geometry. By analyzing how different boundary and source conditions influence the resulting potential fields, we demonstrated that both formulations generate coherent and interpretable maps that encode cell structure and contour information within a single channel. 
Despite using a single output tensor, a single \lone\ term regression loss, and no multi-head predictions or complex post-processing, our models produce potential fields that are accurately segmented by the watershed method. They achieve performance remarkably close and superior to high-end, expensive frameworks such as Cellpose-SAM. Collectively, these results demonstrate that diffusion and Poisson representations offer a 
%mathematically grounded, interpretable, and 
computationally efficient basis for instance-level segmentation in complex microscopy data.